\documentclass[conference]{IEEEtran}
\usepackage[english]{babel}
\usepackage[utf8x]{inputenc}
\usepackage{amsmath}
\usepackage{graphicx}
\usepackage[colorinlistoftodos]{todonotes}
\usepackage{multirow} 
\title{Image-Text Out-Of-Context Detection Using Synthetic Multimodal Misinformation}
\author{
Fatma Shalabi$^{*1,2}$, Huy H. Nguyen$^{*2}$, Hichem Felouat$^{1,2}$, Ching-Chun Chang$^{2}$, and Isao Echizen$^{1,2,3}$ \\
\small{$^{*}$These authors contributed equally} \\
\small{$^{1}$The Graduate University for Advanced Studies, SOKENDAI, Japan} \\
\small{$^{2}$National Institute of Informatics, Japan \ \ \ \ \ \ \ \ $^{3}$The University of Tokyo, Japan} \\
\small{E-mail: \{fatmafaek, nhhuy, hichemfel, cchang, iechizen\}@nii.ac.jp}
}

\begin{document}
\maketitle

\begin{abstract}
Misinformation has become a major challenge in the era of increasing digital information, requiring the development of effective detection methods. We have investigated a novel approach to Out-Of-Context detection (OOCD) that uses synthetic data generation. We created a dataset specifically designed for OOCD and developed an efficient detector for accurate classification. Our experimental findings validate the use of synthetic data generation and demonstrate its efficacy in addressing the data limitations associated with OOCD. The dataset and detector should serve as valuable resources for future research and the development of robust misinformation detection systems.
\end{abstract}

\section{Introduction}
The widespread dissemination of misinformation online, whether in news articles or brief tweets, represents a serious threat to information security and undermines public confidence in online information. The spread of rumors and misinformation can negatively affect people in many ways, such as by being influenced by misleading information, by being victimized by hate speech, by being aggrieved by racism, and by suffering psychological distress~\cite{ali2022combatting}~\cite{mccrae2022multi}. Those who generate online misinformation often utilize multimodal misinformation, such as images and texts, to maximize the effect~\cite{agarwal2019protecting}.

Since misinformation is usually accompanied by images to capture the readers’ attention and add more credibility, a term has emerged to describe the occurrence of an image or text being taken out of its original context or when an image or text is manipulated so as to create misunderstandings and misrepresentations: ``multimodal misinformation''. One common form of multimodal misinformation is Out-of-context (OOC) which refers to separating authentic images and texts from their original context to change (or eliminate) the intended meaning, which can deceive the audience by making the information appear more credible. Despite its prevalence, the use of out-of-context images and texts remains one of the easiest and most effective ways to create realistically appearing misinformation~\cite{aneja2020generalized,wang2022representation}.

Unlike the deepfake generation, the generation of OOC does not require extensive technical knowledge or experience. Moreover, this type of manipulation is harder to detect since neither the image nor the text is distorted. Furthermore, misleading content can be produced in numerous ways~\cite{jaiswal2019aird, shu2017fake}. Existing deepfake detection methods are not helpful since the out-of-context relationship between the image and text is at the semantic level rather than at the signal level~\cite{marra2018detection}. As a result, verifying such content relies heavily on manual fact-checking, which can be costly, time-consuming, and error-prone~\cite{tan2020detecting}. Therefore, there is a pressing need to develop machine learning-based algorithms as an alternative solution for automatically detecting OOC.

Several recent studies on detecting inconsistencies between images and texts have demonstrated promising performance on benchmark datasets. Existing methods have taken various approaches, such as relying on entities, context, or social media responses/reactions to posts, to identify such misinformation~\cite{abdelnabi2022open}. Notwithstanding the advancements made in this field, OOCD still has limitations, including a lack of relevance to real-world situations due to random mismatches in the training and testing datasets, the need for interpretable explanations to verify accuracy and understand the reasoning, the need for scalability to handle the increasing volume of misinformation in real-time, the dependence on specific languages (e.g., English) limiting applicability to other languages, the scarcity of labeled multimodal datasets, and the challenge of detecting and adapting to adversarial attacks based on misinformation. These limitations highlight the need for research and development of more robust and comprehensive methods for OOCD~\cite{huang2022text, luo2021newsclippings}.

In the work reported here, we addressed the challenge of OOCD. \textbf{First}, to address the data limitations, we devised a novel training and evaluation method that uses synthetic data generation to enhance the accuracy of OOCD models. Generating synthetic data, which does not require human labeling effort, expands the diversity and complexity of the training data, enabling the detector to better identify and classify instances of information that deviate from their surrounding context. Furthermore, we can address the problem of explainability by generating a synthetic multimodal dataset that aids in understanding the reasoning behind the detection. \textbf{Second}, we created a novel dataset for OOCD tasks. This dataset should be a valuable resource for future research and benchmarking. \textbf{Finally}, we developed a detector that leverages state-of-the-art machine learning algorithms and techniques to accurately identify OOC multimodal information. This work offers several benefits to the image-text OOCD field by improving the reliability and credibility of image-text data analysis.

%-----------------------------------------------------------------------------------------------------------
\section{Related Work}
\subsection{Available Datasets}
Gathering large-scale labeled multimodal misinformation datasets is difficult due to their scarcity and the substantial manual effort required. Some examples of datasets that have been created for detecting multimodal misinformation include the MultimodAI Information Manipulation (MAIM) dataset~\cite{jaiswal2017multimedia} and the Multimodal Entity Image Repurposing (MEIR) dataset~\cite{sabir2018deep}. The MAIM dataset includes over 239K image-caption pairs with randomly mismatched falsified media. The MEIR dataset introduces swaps over named entities for people, organizations, and locations. It is based on the assumption that, for each image caption (package), there is an unmanipulated related package (geographically near and semantically similar) in the reference set. This enables the integrity of a query package to be verified by retrieving a related package and comparing the two.

Another such dataset is the TamperedNews dataset~\cite{muller2020multimodal} in which named entities specific to people, locations, and events are randomly swapped with other named entities within the article body. These text manipulations can lead to strong linguistic biases. Zellers \textit{et al.}~\cite{zellers2019defending} replaced image captions with synthetic ones that matched the images, which remained relevant to the article’s content. The usefulness of image analysis on this task is rather limited, whereas analyzing the captions and the article body results in the best detection performance. The recently developed COSMOS dataset~\cite{aneja2021cosmos} contains 200K images and 450K captions, where each image is associated with two captions from two sources. The randomly chosen text is used to generate mismatched image-caption pairs. Most of the data in COSMOS is not labeled, although a small subset has been manually annotated as in- or out-of-context.

 Another recently developed dataset is the Visual News dataset~\cite{liu2021visual}, which contains more than one million publicly available news images paired with both captions and news article text collected on a diverse set of topics from news sources in English (The Guardian, the BBC, USA TODAY, and the Washington Post). It was leveraged to create the Visual News Captioner, which generates captions by attending to individual word tokens, named entities in the input news article text, and localized visual features. It was used to create the NewsCLIPpings~\cite{luo2021newsclippings} dataset, a large-scale dataset designed for automatic retrieval of image-text pairs~\cite{liu2021visual}. It contains 988K image-caption pairs for news media mismatch detection. Several strategies can be used to automatically retrieve suitable images for the given captions, including caption-image similarity, caption-caption similarity, person match, and scene match, to identify cases in which there are inconsistent entities or inconsistent semantic context. Finally, the Twitter-COMMs~\cite{biamby2022twitter} dataset consists of multimodal tweets covering such topics as COVID-19, climate change, and military vehicles. It was created using the Twitter API v2~\footnote{https://developer.twitter.com/en/docs/twitter-api/gettingstarted/about-twitter-api}. Hard negatives are created by retrieving the image of the sample tweet with the most textual similarity with a given caption (using the ``Semantics / CLIP Text-Text split'' method of Luo \textit{et al.}~\cite{luo2021newsclippings}), and random negatives are created by selecting an image for a given caption at random to resolve the class imbalance of data collected from Twitter. 

Several studies have focused on human-made fake news detection, resulting in the development of such datasets as FakeNewsNet~\cite{shu2020fakenewsnet} and Fakeddit~\cite{nakamura2020fakeddit}. Another study focused on creating a videos and captions dataset, i.e., a video-based dataset~\cite{mccrae2022multi}, that contains 4,000 real-world Facebook news posts with captions randomly taken from other posts to generate mismatched samples. It was developed for detecting semantic inconsistency between videos and captions in social media posts.
The datasets relevant to the work reported here are summarized in Table \ref{tab:avb_dataset}. The key differences, such as the image-text mismatch procedure used in each dataset, are highlighted.

\begin{table*}[ht]
\centering
\caption{Summary of relevant datasets. Size is the total number of unique samples across all splits.}
\label{tab:avb_dataset}
\resizebox{\textwidth}{!}{%
\begin{tabular}{l|l|l|l|l|l}
\hline
\textbf{Dataset} & \textbf{Year} & \textbf{Data} & \textbf{Source} & \textbf{Mismatch} & \textbf{Size} \\ \hline
{MAIM~\cite{jaiswal2017multimedia}} &
  {2017} &
  {Caption, Image} &
  {Flickr} &
  Random &
  {239k} \\
{MEIR~\cite{sabir2018deep}} &
  {2018} &
  {Caption, Image, GPS} &
  {Flickr} &
  Text entity manipulation &
  {57k} \\
{TamperedNews~\cite{muller2020multimodal}} &
  {2020} &
  {Article, Image} &
  {BreakingNews} &
  Text entity manipulation &
  {776k} \\
  {VisualNews~\cite{liu2021visual}} &
  {2020} &
  {Article, Caption, Image, Metadata} &
  {4 news agencies: Guardian, BBC, USA Today, Washington Post} &
  / &
  {1.25M} \\
{COSMOS~\cite{aneja2021cosmos}} &
  {2021} &
  {Caption,   Image} &
  {News Outlets} &
  Two sources (3k labeled) &
  {453k} \\
{NewsCLIPpings~\cite{luo2021newsclippings}} &
  {2021} &
  {Caption,   Image} &
  {VisualNews} &
  Automatic retrieval &
  {988k} \\
{Twitter-COMMs~\cite{biamby2022twitter}} &
  {2022} &
  {Caption, Image} &
  {Tweets: COVID-19, Climate  Change, Military Vehicles} &
  Random \& Automatic retrieval &
  {884k} \\ \hline
\end{tabular}%
}
\end{table*}

%-----------------------------------------------------------------------------------------------------------
%-----------------------------------------------------------------------------------------------------------
\subsection{Image Captioning}
Generating a high-quality description of an image using natural language is the goal of image captioning. This is a challenging task for both humans and machines~\cite{lin2014microsoft} ~\cite{sharma2018conceptual}. However, recent advancements in large-scale vision-language models (VLMs) and large-language models (LLMs) have made it possible for models to generate captions that are increasingly difficult to distinguish from those generated by humans. Some models, such as OFA~\cite{wang2022ofa}, Flamingo~\cite{alayrac2022flamingo}, and BLIP-2~\cite{li2023blip}, have demonstrated impressive performance in image captioning tasks. In certain evaluations, several approaches have been rated as good as or better than human users.

Researchers have explored various approaches, such as ``Plug-and-play VQA'' (visual question answering)~\cite{tiong2022plug}, and ``PromptCap'' (Prompt-guided image Captioning)~\cite{hu2022promptcap}, to generating natural language prompts in conjunction with a generative pre-trained transformer (GPT) to achieve state-of-the-art (SOTA) performance in visual question-answering. Mokady \textit{et al.}~\cite{mokady2021clipcap} utilized continuous embedding as a prompt for the GPT-2 model~\cite{radford2019language} to achieve strong single-viewpoint image captioning performance. Zeng \textit{et al.}~\cite{zeng2022socratic} used a CLIP-based model~\cite{radford2021learning} to extract key tags from an image and then used the GPT-3 model~\cite{brown2020language} with a specialized prompt to generate an appropriate image caption.

%-----------------------------------------------------------------------------------------------------------
%-----------------------------------------------------------------------------------------------------------
\subsection{Text-to-Image Generation}
Deep generative models have achieved tremendous success in text-to-image (T2I) generation tasks and have recently attracted intensive attention. Existing T2I generation models can be categorized as GAN-based, VAE-based, and diffusion-based. Although GAN-based~\cite{creswell2018generative} and VAE-based models~\cite{kingma2014auto} synthesize images with promising quality and diversity, they still cannot match user descriptions well.

Diffusion-based models~\cite{ho2020denoising} have demonstrated unprecedentedly high-quality and controllable image generation and have been broadly applied to T2I generation tasks. They were first introduced into T2I generation with classifier-free guidance. The introduction of the DALL-E deep learning model~\cite{ramesh2021zero} made it possible to perform zero-shot T2I generation with arbitrary text input. It quickly inspired several follow-up models, including Glide~\cite{nichol2022glide}, a text-guided diffusion model supporting both image generation and editing. It implements the text generation feature in the transformer blocks used in the denoising process. The CogView model~\cite{ding2021cogview} is a transformer with a vector quantized-variational autoencoder~\cite{van2017neural} tokenizer similar to that in DALL-E~\cite{ramesh2021zero}, and it supports Chinese text input. The Imagen model~\cite{saharia2022photorealistic} has a T2I structure that does not use latent images and instead directly diffuses pixels using a pyramid structure. The Stable Diffusion model~\cite{dhariwal2021diffusion} is a large-scale implementation of the latent diffusion model (LDM)~\cite{rombach2022high} that achieves T2I generation. These models use pre-trained large-scale text encoders to provide more controllable guidance signals. They either introduce super-resolution models or encode images into the low-dimensional latent space to achieve a trade-off between efficiency and high resolution. 

%-----------------------------------------------------------------------------------------------------------
%-----------------------------------------------------------------------------------------------------------
\subsection{Multimodal Misinformation Detection }
Studies on MMD have focused on OOC image-text pairs or cross-modal named entity inconsistencies. Aneja \textit{et al.}~\cite{aneja2021cosmos} proposed using a self-supervised approach to establish whether two captions accompanying an image are consistent. However, the unlabeled dataset they collected does not support veracity detection training and evaluation. Furthermore, the assumptions made about the availability of such a dataset may affect the model’s feasibility in a generic test setting, and its disproportionate reliance on text mode may lead to an increased decision-level bias.

Luo \textit{et al.}~\cite{luo2021newsclippings} leveraged the expressiveness of a large pre-trained contrastive model, CLIP~\cite{radford2021learning}, to classify mismatches on the basis of retrieval; in addition, they evaluated two VLMs, CLIP and VisualBERT~\cite{li2019visualbert}, on the proposed dataset and achieved classification accuracies of 60.23\% and 54.81\%, respectively. They showed that both machine and human detection are limited, indicating that the task is challenging. Abdelnabi \textit{et al.}~\cite{abdelnabi2022open} collected evidence for both visual and textual components for use in performing cycle consistency checks. However, relying on such external information for validity checks makes the entire approach very expensive, complex, memory intensive, and challenging to deploy in a generic test setting. Furthermore, in this use case setting, the primary objective was to verify the content of a news item quickly. Therefore, sufficient reliable information related to the news topic may not be available yet on the web.

Huang \textit{et al.}~\cite{huang2022text} used two VLMs, CLIP and VinVL~\cite{zhang2021vinvl}, for the multimedia inconsistency detection task. They encoded each image and its corresponding caption separately and then compared their embeddings; dissimilar embeddings suggested an out-of-context text-image pair. They evaluated their method on several large-scale datasets. However, their method does not address the potential biases in the VLMs used, which may have affected the accuracy of the OOCD. In addition, their method relies on the availability of textual information associated with the image, which may not always be available. The self-supervised distilled learner (SSDL) method devised by Mu \textit{et al.}~\cite{mu2023self} uses two strategies to estimate the cross-modal consistency status in terms of a binary class label (Pristine or Falsified). However, the effectiveness of the SSDL approach relies heavily on the quality and accuracy of the annotated data used during the training phase. Errors and biases in the annotations can severely degrade the model's ability to learn and generalize effectively. Moreover, the SSDL approach fails to address potential biases or provide strategies for mitigating them. Furthermore, the training and fine-tuning of the self-supervised models are computationally intensive and take a considerable amount of time.

The method devised by Zhang \textit{et al.}~\cite{zhang2023detecting} detects OOC multimodal misinformation by using an interpretable neural-symbolic model. However, it does not address other types of misinformation and lacks empirical evaluation and comparison to other approaches. The model shows theoretical promise, but without thorough experimental validation, its effectiveness and superiority over other methods are uncertain. It is also subject to challenges like scalability, efficiency, and generalization across domains and languages, raising concerns about its practical viability. Table \ref{tab:ext_methods} summarizes existing methods for detecting multimodal misinformation.

\begin{table*}[ht]
\centering
\caption{Summary of existing multimodal misinformation detection methods.}
\label{tab:ext_methods}
\resizebox{\textwidth}{!}{%
\begin{tabular}{l|l|l|l|ll}
\hline
\textbf{Work} & \textbf{Year} & \textbf{Models} & \textbf{Dataset} & \textbf{Method Description} \\ \hline
Aneja \textit{et al.}~\cite{aneja2021cosmos} &
 2021 &
 \begin{tabular}[c]{@{}l@{}}spaCy NER, Mask-RCNN, Object Encoder,\\ Universal Sentence Encoder, SBERT\end{tabular} &
 COSMOS &
 \begin{tabular}[c]{@{}l@{}}Uses a self-supervised approach to establish whether two captions \\accompanying an image are consistent.\end{tabular} \\ \hline
Luo \textit{et al.}~\cite{luo2021newsclippings} &
 2021 &
 CLIP, VisualBERT &
 VisualNews &
 \begin{tabular}[c]{@{}l@{}}Classifies mismatches on the basis of retrieval (CLIP and\\ VisualBERT were evaluated on proposed dataset).\end{tabular} \\ \hline
Abdelnabi \textit{et al.}~\cite{abdelnabi2022open} &
 2022 &
 \begin{tabular}[c]{@{}l@{}}ResNet152, sentence transformer,\\ BERT + LSTM ,spaCy NER, CLIP\end{tabular} &
 NewsCLIPpings &
 \begin{tabular}[c]{@{}l@{}}Collects evidence for both visual and textual components for use \\in cycle consistency checks.\end{tabular} \\ \hline
Huang \textit{et al.}~\cite{huang2022text} &
 2022 &
 CLIP, VinVL &
 \begin{tabular}[c]{@{}l@{}}NewsCLIPpings\\\& private datasets \\(DARPA)\end{tabular} &
 \begin{tabular}[c]{@{}l@{}}Encodes an image and its corresponding caption separately and then\\ compares their embeddings for multimedia inconsistency detection task.\end{tabular} \\ \hline
Mu \textit{et al.}~\cite{mu2023self} &
 2023 &
 CLIP, SimCLRV2 , ResNet50 &
 NewsCLIPpings &
 \begin{tabular}[c]{@{}l@{}} Uses two networks, Student Classifier (semi-supervised learning)\\ and Teacher Network (self-supervised learning), for SSDL.\end{tabular} \\ \hline
Zhang \textit{et al.}~\cite{zhang2023detecting} &
 2023 &
 \begin{tabular}[c]{@{}l@{}}MLPs and SAFE, VisualBERT, CLIP, VINVL, \\FaceNet+BERT\end{tabular} &
 NewsCLIPpings &
 \begin{tabular}[c]{@{}l@{}} Uses a hybrid model that combines a neural network and symbolic reasoning and \\captures context and semantics of content by using textual and visual information. \end{tabular} \\ \hline
\end{tabular}%
}
\end{table*}

%-----------------------------------------------------------------------------------------------------------

%-----------------------------------------------------------------------------------------------------------
\section{Proposed Approach}
Our proposed approach has distinctive features. Unlike existing approaches, it achieves remarkable results without relying on external sources or additional data to enhance and augment the learning process of the classifier. Furthermore, it eliminates the need to conduct Internet-based searches to validate the accuracy of the classifier's decision or to establish the connection between an image and the corresponding caption. These features will allow us to advance our approach further as the underlying models evolve.

%-----------------------------------------------------------------------------------------------------------
%-----------------------------------------------------------------------------------------------------------
\subsection{Problem Statement}
Detecting OOC is a challenging problem that not only stems from multimodal semantic mismatches but also from the lack of mature problem definitions and datasets for evaluation. Deep neural networks require large amounts of data, and manual annotation is labor-intensive. Furthermore, the limited evaluation of real-world misinformation and the absence of comparisons with synthetic multimodal misinformation make it difficult to assess progress in the field.
One of the main reasons for the spread of misinformation is the increasing use of generative AI to create multimodal misinformation. However, few methods exist to detect misinformation because annotated datasets are scarce. Our proposed approach is aimed at detecting OOC by using synthetic misinformation generated from the original dataset. This maximizes the utilization of limited annotated data by generating additional samples. We estimate the level of consistency between the multimodal data using the generated synthetic data, enabling us to assess the overall consistency within the original data. This estimation is performed as a binary classification (match or mismatch).

%-----------------------------------------------------------------------------------------------------------
%-----------------------------------------------------------------------------------------------------------
\subsection{Data Preparation}
\begin{figure}[ht]
\begin{center}
    \includegraphics[scale=0.4]{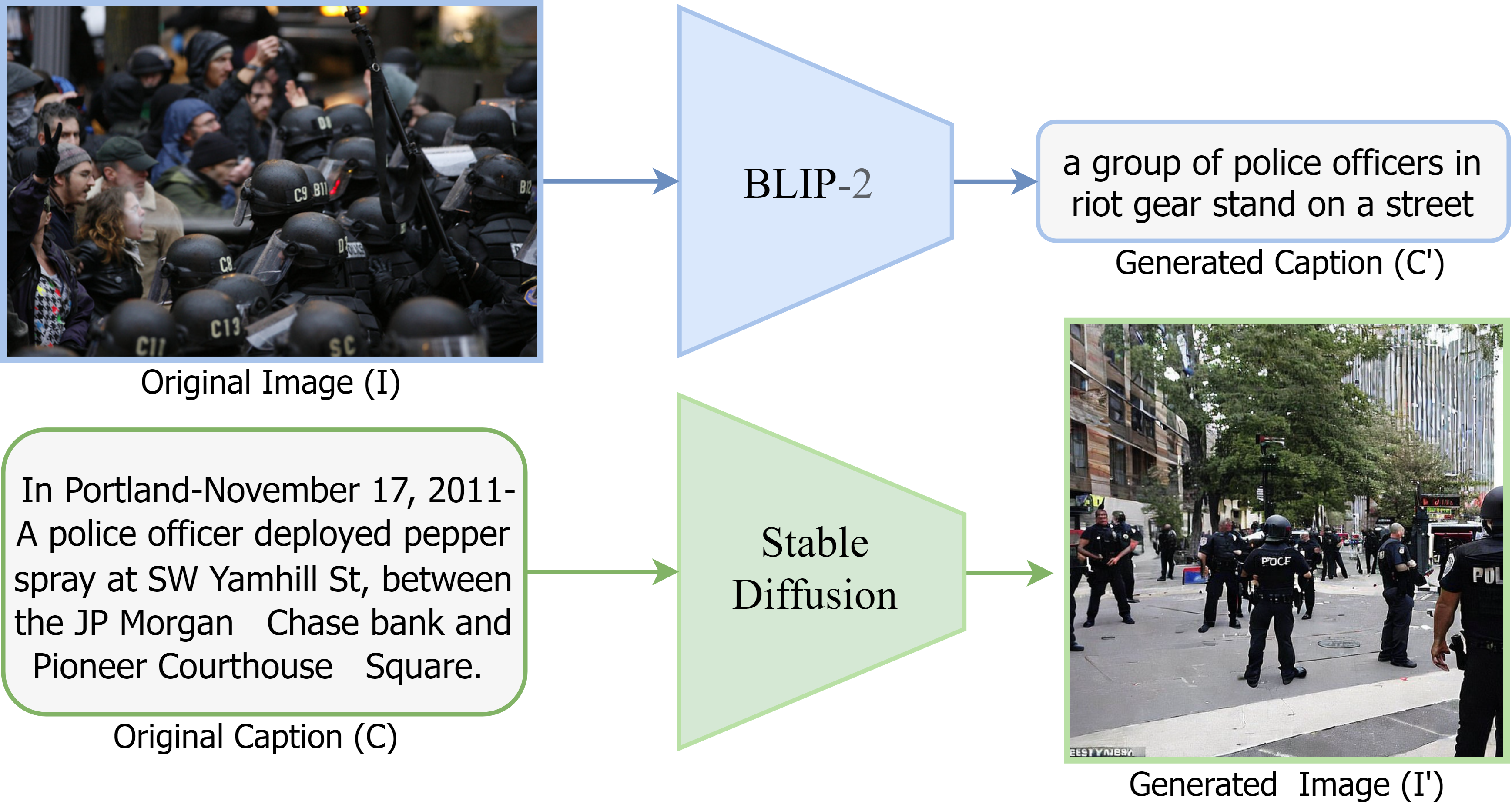}
    \caption{Representation of steps involved in dataset preparation.
    }
    \label{fig:pre}
\end{center}
\end{figure}

Our approach involves the creation of synthetic data versions using the original image, caption, and label from the original dataset\cite{luo2021newsclippings}. This process consists of two key steps. Firstly, Starting with the original image \(I\), we utilize an image captioning model to generate a corresponding caption \(C{'}\). Subsequently, employing a text-to-image T2I model, we generate an image \(I{'}\) based on the original caption \(C\). As a result, our augmented dataset comprises the original images accompanied by their corresponding captions and the generated images and generated captions.  The dataset preparation process is outlined in Figure \ref{fig:pre}.

%-----------------------------------------------------------------------------------------------------------
%-----------------------------------------------------------------------------------------------------------
\subsubsection{Caption Generation}
The process of generating a caption for the original image is performed using a pre-trained image captioning model to generate multiple captions based on a given image. The captions produced are condensed into a single (summary) caption to create the output. Formally, given an image $I \in N^{H\times W\times C}$, we aim to make a sequence of \textit{m} tokens $x_{1}...x_{m}$ describing the image in some vocabulary \textit{V} and produce a probability distribution $P(x_{m} \in V|I, x_{1} ... x_{m-1})$, which is the probability of the next token in the sentence, given both the image and all the previous tokens.

We used the BLIP-2~\cite{li2023blip} model and fine-tuned it specifically for the image captioning task. BLIP-2 is a vision-language pre-training framework designed to utilize large amounts of noisy web data for effective pre-training. The model uses synthetic image-caption pairs from a seed captioning model and a filter to eliminate low-quality synthetic captions, resulting in a large dataset. BLIP-2 has shown remarkable transfer performance in many vision-language tasks, and it excels in zero-shot and fine-tuning scenarios for image captioning because the BLIP-2 model is trained on massive image and language datasets (the MS-COCO datasets~\cite{lin2014microsoft}). 

\begin{figure*}[ht]
    \begin{center}
    \includegraphics[width=17cm]{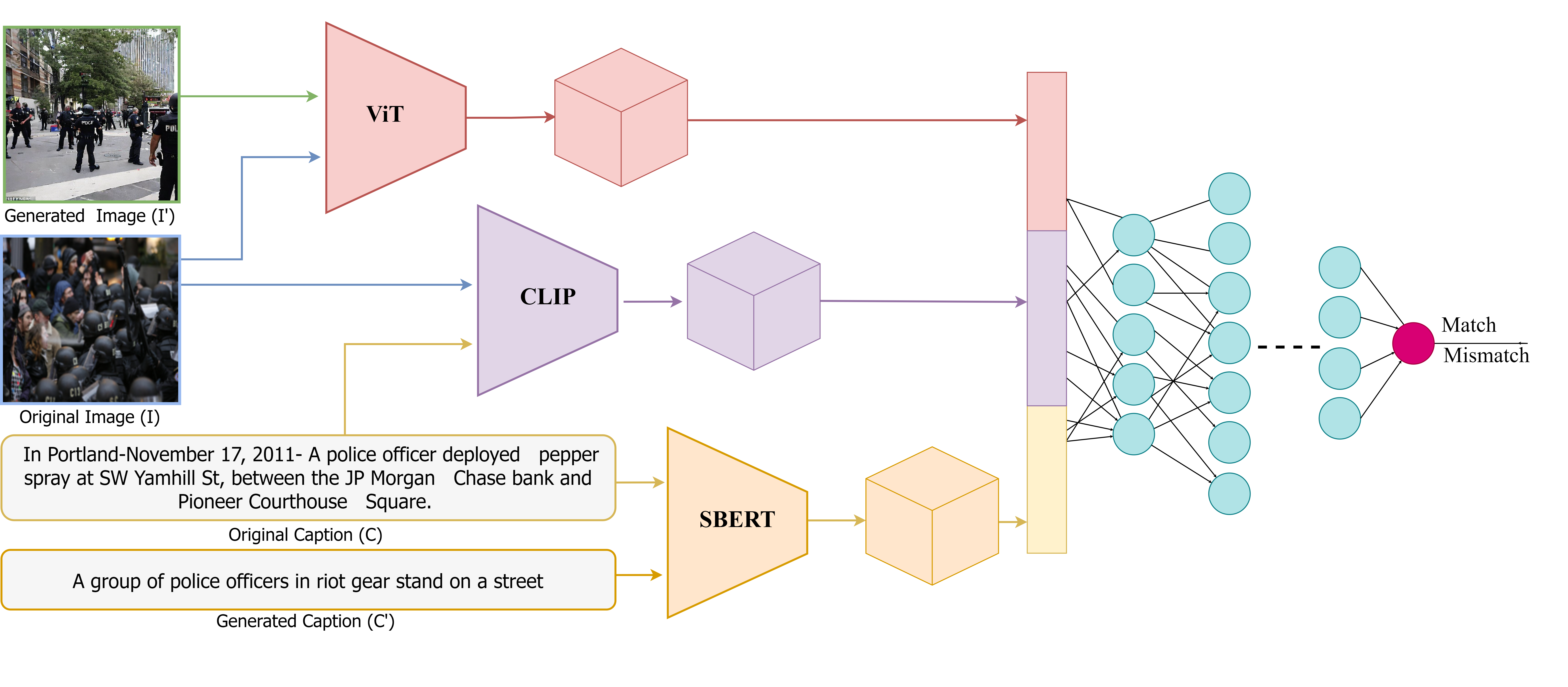}
    \caption{Overview of our proposed approach.
    }
    \label{fig:app_0}
    \end{center}
\end{figure*}

%-----------------------------------------------------------------------------------------------------------
%-----------------------------------------------------------------------------------------------------------
\subsubsection{Image Generation}
Our work uses an LDM~\cite{rombach2022high} loaded with Stable Diffusion v1.4 weights to generate synthetic images. The Stable Diffusion model was trained on a subset of the LAION5B dataset~\cite{cherti2023reproducible} and generated $(512\times512)$ pixel images, which are then resized to the native image size of each dataset. To provide conditioning from text prompts, a frozen CLIP ViT-L/14 text encoder is used in conjunction with Stable Diffusion~\cite{rombach2022high,saharia2022photorealistic}. Although LDMs can generate images from several different conditionings (image, text, semantic map), we only generated images using text prompts (captions). This enables zero-shot learning; it requires only a text prompt describing the contents of the image without any real images. 

To generate images through text prompts, we require a prompt (caption) that describes the desired image's contents to guide the diffusion process. The prompt is first projected into an intermediate representation via CLIP's text encoder and then mapped to the intermediate layers of the LDM's denoising UNet using cross-attention~\cite{rombach2022high}. This guidance helps the LDM diffuse a latent representation of an image starting from Gaussian noise. The resulting latent representation is then decoded back into the pixel domain to produce the final image.

Three primary hyperparameters control the generation process. One is the number of steps the denoising diffusion implicit model (DDIM) takes in the denoising process. More steps generally result in more realistic and coherent images, while fewer steps result in more disjointed and surreal images~\cite{ho2020denoising}. We generated synthetic images using 500 DDIM steps. The second hyperparameter is the unconditional guidance scale (UGC), which controls the scale between the precision of the generated image matching, the provided prompt, and generation diversity. This is accomplished by scaling between the jointly trained conditional and unconditional diffusion models~\cite{ho2022classifier}. A lower UGC value means less guidance and more diversity, while a higher value means more guidance and less diversity. The third hyperparameter is the Seed, which is the starting point for diffusion, generating the initial Gaussian.

%-----------------------------------------------------------------------------------------------------------
%-----------------------------------------------------------------------------------------------------------
\subsection{Out-Of-Context Detection Model}
Our proposed multimodal model, which effectively detects OOC by leveraging synthetic misinformation, is illustrated in Figure~\ref{fig:app_0}. Deep features are extracted from our dataset (\(I\), \(C\), \(I{'}\), \(C{'}\)) and used by the classifier to more effectively train the model and to enhance its ability to discern patterns of consistency between pristine and falsified examples. These features are extracted from the original image \(I\) and caption \(C\) by using a pre-trained multimodal CLIP model~\cite{radford2021learning}. Additionally, features are extracted from the original caption \(C\) and the generated caption \(C{'}\) by using Sentence-BERT~\cite{reimers2019sentence}. A vision transformer (ViT)~\cite{dosovitskiy2021image} is used to extract features from the original image \(I\) and the generated image \(I{'}\). All of these features are concatenated and input into the classifier.

The output of the classification process yields one of two scenarios depending on the contextual relationship between the original image \(I\) and its corresponding caption \(C\). In the first scenario, if the original image and caption \((I, C)\) are within the same context, the generated image \(I{'}\) derived from the original caption will inherently share the context of the original caption \(C\) and align with the original image \(I\). Similarly, the generated caption \(C{'}\) will be in the same context as the original image \(I\), which is consistent with the original caption \(C\). As a result, there is a high similarity between the original image and the generated image \((I, I{'})\), and between the original caption and the generated caption \((C, C{'})\).

If the original image and caption \((I, C)\) are out of context, the original image and the generated image \(I, I{'})\) and also the original caption and the generated caption \((C, C{'})\) will have low similarity scores, as follows:

 \begin{gather*}
 ((I,C,y)= \begin{cases} If \quad y = Match &\\  
  High_{Similarity}((I,C),(C,C^{'}),(I,I^{'})) &\\ 
  &\\ Otherwise \quad y = Mismatch &\\  
  Low_{Similarity}((I,C),(C,C^{'}),(I,I^{'}))  
 \end{cases}
 \label{gather:1}
 \end{gather*}\\
where y denotes the label of the pair of the original image and caption (match or mismatch). \(High_{Similarity}\) is defined as the highest similarity score among all pairs, indicating a high correlation between the image and caption. Conversely, \(Low_{Similarity}\) is defined as the minimum similarity score among all pairs, indicating a low correlation between the image and caption.
%-----------------------------------------------------------------------------------------------------------
%-----------------------------------------------------------------------------------------------------------
\section{Experimental and Results}
In this section, we describe our experimental setting and then present and analyze the results. 

%-----------------------------------------------------------------------------------------------------------
%-----------------------------------------------------------------------------------------------------------   
\subsection{Dataset}
We used the NewsCLIPpings~\cite{luo2021newsclippings} dataset, which consists of pristine and falsified OOC images. It was constructed on the basis of the VisualNews~\cite{liu2021visual} corpus, which includes news articles from four major news outlets: the BBC, The Guardian, the Washington Post, and USA Today. The NewsCLIPpings dataset was constructed using various methods for automatically retrieving convincing images for a given caption, capturing cases with inconsistent entities or semantic context (such as text-text similarity or image-image similarity), etc. We use the Merged/Balanced dataset, which contains data representative of all retrieval methods and contains 85K balanced pristine and falsified examples, 71,072 training examples, 7,024 validation examples, and 7,264 test examples, divided into a 10:1:1 ratio. To initiate our generation-assisted detection approach, we used the original image-caption pairs as prompts to generate synthetic image-caption pairs, as illustrated in Figure \ref{fig:pre}.
%-----------------------------------------------------------------------------------------------------------
%-----------------------------------------------------------------------------------------------------------
\subsection{Experimental Settings}
We conducted two evaluation experiments to assess the merits of our proposed approach comprehensively. To establish the effectiveness and efficacy of our approach, we compared the results with those of previous studies. This comparative analysis serves as a robust validation of our proposed approach.

\subsubsection{Experiment using Cosine Similarity}
In the first experiment, we used CLIP (ViT-B/32)~\cite{radford2021learning}, a powerful model, to extract embedded features from the original image-caption pairs. This process yielded a 512-dimensional vector, represented as \((I_{embdCLIP}, C_{embdCLIP})\), for each pair. To quantify the similarity between the original image and caption \((I, C)\), we used the cosine similarity metric to calculate \(CLIP_{similarity}\). To evaluate the similarity between the original caption and the generated caption \((C, C^{'})\), we used Sentence-BERT~\cite{reimers2019sentence} (all-mpnet-base-v2) for embedding both captions and obtained a 768-dimensional vector \((C_{embdSBERT}, C^{'}_{embdSBERT})\) for each pair. We then measured the similarity between their embeddings by using the cosine similarity metric, resulting in \(SBERT_{similarity}\).

Furthermore, we used a vision transformer (ViT\_L\_16)~\cite{dosovitskiy2021image} to extract the embeddings from both the original image and the generated image \((I, I^{'})\); this embedding yielded a 1,024 item vector, represented as \((I_{embdViT}, I^{'}_{embdViT})\), for each image. Using cosine similarity, we compared the embeddings of \((I , I^{'})\) to obtain \(ViT_{similarity}\), thereby quantifying their similarity. Next, the three similarity values were fed into machine learning classifiers to distinguish patterns between the \((I, C)\) pairs with coherent context and those with incoherent context. Table \ref{tab:exp1} presents the experimental results in comparison with those for four other classifiers.

\subsubsection{Experiment using Feature Extraction}
In the second experiment, we used feature extraction techniques to analyze the effectiveness of different feature map combinations. The feature maps were obtained from three models (see Figure 2), namely CLIP, Sentence-BERT, and Vision Transformer, as represented by \((I_{embdCLIP}, C_{embdCLIP})\), \((C_{embdSBERT}, C^{'}_{embdSBERT})\), and \((I_{embdViT}, I^{'}_{embdViT})\), respectively.

To analyze the feature maps, we followed a two-step approach. We first flattened each pair of feature maps and separately fed them into machine learning classifiers and a transformer designed for tabular data analysis. Next, we consolidated all the feature maps and fed them into both the machine learning classifiers and the transformer. To optimize the classification process, we utilized dimensionality reduction techniques on the combined set of concatenated features. This enabled us to reduce the complexity of the data while retaining important information. To thoroughly investigate the potential of different feature map combinations, we systematically evaluated multiple combinations. The results of these evaluations are summarized in Table \ref{tab:exp2}.

%-----------------------------------------------------------------------------------------------------------

\begin{table}[ht]
\centering
\caption{Detection performance (in \%) based on similarity scores }
\label{tab:exp1}
\resizebox{\columnwidth}{!}{%
\begin{tabular}{l|cccccc}
\hline
\multirow{3}{*}{\textbf{Model}} & \multicolumn{6}{c}{\textbf{Similarity}}                                     \\ \cline{2-7} 
 & \multicolumn{2}{c|}{CLIP+SBERT} & \multicolumn{2}{c|}{CLIP + ViT} & \multicolumn{2}{c}{CLIP+SBERT+ViT} \\ \cline{2-7} 
                                & ACC & \multicolumn{1}{c|}{AUC} & ACC & \multicolumn{1}{c|}{AUC} & ACC & AUC \\ \hline
SVM                             & 66  & \multicolumn{1}{c|}{72}    & 65  & \multicolumn{1}{c|}{72}    &  {67}   & {73}    \\
Random Forest                   & 62  & \multicolumn{1}{c|}{71}    & 62  & \multicolumn{1}{c|}{67}    &  {67}   & {72} \\
XGBoost                         & 66  & \multicolumn{1}{c|}{72}    & 64  & \multicolumn{1}{c|}{72}    &  {67}   & {73}  \\ \hline
\textbf{MLP}                    & 67  & \multicolumn{1}{c|}{73}    & 66  & \multicolumn{1}{c|}{73}    &  {\textbf{68}} & {\textbf{74}}  \\ \hline
Transformer                     & 66  & \multicolumn{1}{c|}{72}    & 65  & \multicolumn{1}{c|}{72}    &  {67}   & {73}  \\ \hline
\end{tabular}%
}
\end{table}
%--------------------------------------------------------------------------------
\begin{table}[ht]
\centering
\caption{Detection accuracy (in \%)  based on Feature maps }
\label{tab:exp2}
%\resizebox{\columnwidth}{!}{%
\begin{tabular}{l|c|c|c}
\hline
\multirow{2}{*}{\textbf{Model}} & \multicolumn{3}{c}{\textbf{Features}}                  \\ \cline{2-4} 
                     & \multicolumn{1}{l|}{CLIP}        & \multicolumn{1}{l|}{CLIP+SBERT}  & CLIP+SBERT+ViT \\ \hline
SVM                    & 50 & 50 & 54 \\
Random Forest          & 49 & 50 & 52 \\
XGBoost                & 52 & 53 & 55 \\
MLP                    & 53 & 55 & 56 \\
\hline
\textbf{Transformer} & \textbf{56} & \textbf{60} & \textbf{62}    \\ \hline
\end{tabular}%
%}
\end{table}

%***************************************************************************************************

%-----------------------------------------------------------------------------------------------------------
\subsection{Discussion}
OOCD presents substantial complexities due to the unaltered nature of the image and its accompanying caption. These two elements exist in separate contexts, often featuring the same names of individuals and places but with different temporal attributes, such as dates. This disparity between the image and its caption poses a considerable challenge in detecting the asymmetry, necessitating further efforts to develop robust models in the future. Despite these complexities, the results obtained with our proposed approach are highly encouraging, motivating the continued advancement of this approach.

Our proposed approach resulted in the highest accuracy among the approaches compared, surpassing other research in this field, as shown in Table \ref{tab:comp}. This table compares our best experimental accuracy with those of previous studies using the same dataset without leveraging additional information to enhance the classification process. Further development of this approach is expected to yield even better results as future research introduces more powerful models capable of accurately measuring the similarity between images or captions and providing improved image captioning.

The efficiency of our approach makes it potentially suitable for real-time operation, delivering prompt responses. This efficiency arises from the parallel implementation of the models used. Furthermore, alongside the OOCD model, we built a dataset of generated images from real captions that researchers in this and related fields can utilize as valuable resources in their work.

\begin{table}[ht]
\centering
\caption{Comparison Results of Our Proposed Approach with Other Approaches using the Same Dataset.}
\label{tab:comp}
\begin{tabular}{l|c|c} 
\hline
\textbf{Paper} & \textbf{Year} & \textbf{Accuracy} (\%) \\ 
\hline
Luo \textit{et al.}~\cite{luo2021newsclippings} & 2021 & 65.9 \\
Huang \textit{et al.}~\cite{huang2022text} & 2022 & 65.2 \\
Abdelnabi \textit{et al.}~\cite{abdelnabi2022open} & 2022 & 66.1 \\
Zhang \textit{et al.}~\cite{zhang2023detecting} & 2023 & 62.8 \\ \hline
\textbf{Ours} & 2023 & \textbf{68.0} \\
\hline
\end{tabular}
\end{table}
%-----------------------------------------------------------------------------------------------------------
\section{Conclusion}
Our proposed approach to Out-Of-Context detection uses synthetic data generation. The dataset and detector we developed enable the accurate identification of misinformation across various modalities. The results of extensive experimentation and evaluation validate the effectiveness of the proposed approach, with which a promising classification accuracy rate of 68\% was achieved for misinformation detection. The dataset and detector should prove valuable for future research and facilitate the development of robust misinformation detection systems.
%-----------------------------------------------------------------------------------------------------------
%-----------------------------------------------------------------------------------------------------------
\section{Acknowledgment}
This work was partially supported by JSPS KAKENHI Grants JP18H04120, JP20K23355, JP21H04907, and JP21K18023, and by JST CREST Grants JPMJCR18A6 and JPMJCR20D3, Japan.

%-----------------------------------------------------------------------------------------------------------

%-----------------------------------------------------------------------------------------------------------
%-----------------------------------------------------------------------------------------------------------
\end{document}